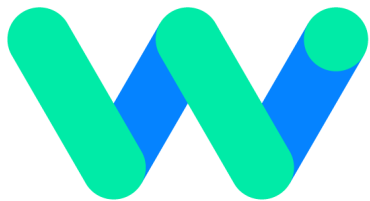

# Waymo's Safety Methodologies
# and Safety Readiness Determinations

October 2020

# Executive Summary

As the world's most experienced developer of automated driving systems ("ADSs"), Waymo has extensive experience in developing and applying state-of-the-art safety methodologies.[1] Waymo's methodologies help implement Waymo's forward-looking safety philosophy: *Waymo will reduce traffic injuries and fatalities by driving safely and responsibly, and will carefully manage risk as we scale our operations.*

Waymo's safety methodologies, which draw on well established engineering processes and address new safety challenges specific to Automated Vehicle ("AV") technology, provide a firm foundation for safe deployment of our Level 4 ADS, which we also refer to as the Waymo Driver™.[2] Waymo's determination of its readiness to deploy its AVs safely in different settings rests on that firm foundation and on a thorough analysis of risks specific to a particular Operational Design Domain ("ODD").

Waymo's Current Operations. Waymo currently operates our AVs primarily in California and Arizona, with additional testing in several other states including Michigan,Texas, Florida and Washington. In Metro Phoenix, our fleet of Chrysler Pacifica AVs has been transporting passengers in various types of self-driving (i.e., ADS-operated) service since 2017, including a driverless service. To date, Waymo has compiled over 20 million self-driving miles on public roads operating in over 25 cities, including 74,000 driverless miles.

Approaches to Assessing Automated Vehicle Safety. Various private and governmental organizations have proposed a wide range of methodologies for measuring or demonstrating AV safety. Several common themes emerge from these various approaches, including: simulation-based testing is an extremely important part of development and safety validation of an ADS; test scenario selection should be tailored for risks likely to occur in a given ODD; and, as in all transportation, transportation in an ADS-equipped vehicle involves some level of risk, and a guiding principle should be to avoid risks that are unreasonable.

Waymo's Safety Methodologies. Waymo's safety methodologies focus on the development, qualification, deployment and sustained field operation of a unique product: a Level 4 ADS that can perform the entire Dynamic Driving Task ("DDT"), with no human driver present, in both normal traffic conditions and very challenging scenarios that we have reason to expect may occur in a specific ODD. While we learn from certain widely accepted engineering processes and principles, we tailor them--as informed by our extensive experience in Level 4 technology--for this purpose. We continuously refine those methodologies in an incremental way as we scale our operations. Waymo's various safety methodologies are supported by three basic types of system-level testing (simulation, closed-course, and public road), which are supplemented by various forms of component and subsystem testing. These types of testing are in constant interaction; each informs and complements the other. We describe the essence of our methodologies under headings that refer to the three layers of our technology: hardware, ADS behavior, and vehicle operations.

---

[1] The primary authors of this overview of Waymo's safety methodologies are Nick Webb and Daniel Smith, with considerable contributions from others including, but not limited to: Christopher Ludwick, Trent Victor, Qi Hommes, Francesca Favarò, George Ivanov, and Tom Daniel. Many others across Waymo contributed to Waymo's Safety Framework, which embodies these methodologies.
[2] Generally, Waymo refers to its ADS as the "Waymo Driver" in order to convey the message that our ADS is designed to handle the entire dynamic driving task.

*Hardware layer*. We begin by purchasing safe, fully certified vehicles from experienced vehicle manufacturers. As part of development of our base vehicles we specify the inclusion of redundant braking and steering actuators, which we feel is necessary for safe, driverless vehicles. The performance and fault tolerance of these motion control actuators is dictated by a thorough set of technical requirements, specifying performance in both nominal and faulted conditions. These requirements are backed by extensive verification including hardware-in-the-loop and closed course testing, as well as validated on closed courses and in the field.

To the base vehicle and motion control systems, Waymo adds our own sensing systems, including a multitude of lidar, radar, camera, inertial and audio units that provide an expansive understanding of the driving environment. These sensing systems are designed to meet rigorously defined performance and safety requirements. To run our advanced behavioral software, Waymo has developed a state-of-the-art computational platform that combines extreme performance with proven reliability and fault tolerance. We have designed our system to have a portfolio of fault responses tailored and matched to any failure. Waymo has also developed a robust process to identify, prioritize, and mitigate cybersecurity threats.

*Behavioral Layer.* The behavioral layer describes the software that is capable of directing safe driverless movement of our AVs on public roads. There are three primary capabilities on which we evaluate the performance of the ADS's behavioral layer: avoidance of crashes, completion of trips in driverless mode, and adherence to applicable driving rules. Our approach begins with hazard analysis, by which robustness is built into our designs from the beginning. We then heavily leverage scenario-based verification, to ensure that the ADS behavior is in line with our requirements and expectations. Finally, we subject our system to large scale simulated deployments (either through large scale log playback or public roads operations with counterfactual simulations after vehicle operator dis-engage) which allow us to empirically measure aggregate performance metrics.

*Operations Layer.* Waymo's safety program ensures application of industry-leading safety practices in the operation of our AVs, such as a fatigue management program for our trained vehicle operators, incident response planning and preparation, and coordination with law enforcement and emergency responders on how to deal safely with driverless vehicles. Waymo also recognizes the importance of seat belt use in any vehicle and we take a number of steps to encourage our AV passengers to use their belts. We have a fleet response team that can provide remote assistance to the ADS if needed. Waymo's Risk Management Program identifies, prioritizes, and drives the resolution of potential safety issues before new or updated features or software are used on public roads or tested at our structured test facility. Our Field Safety Program identifies and effectuates appropriate disposition of potential safety issues based on information collected after updated features or software have been released for driving on public roads. Consistent with Waymo's strong safety culture, the field safety process collects and helps resolve potential safety concerns from many other sources, including employees, our riders, the public or suppliers.

Safety Governance. Waymo's governance process includes a tiered system of analyzing safety issues that arise from the field safety process, risk management, impending deployment decisions, or any other source. The Waymo Safety Board brings together executive leaders from our Safety, Engineering and Product teams to resolve these safety issues, approve new safety activities, and ensure our entire safety framework is kept current.

Waymo's Readiness Determinations. At certain points in time Waymo needs to make a discrete determination resting on its safety methodologies with regard to the readiness of a specific configuration of the ADS for a specific deployment. The determination is focused on the ODD, the specific use case, and the particular vehicle platform. All of Waymo's operations--closed course or public roads, with or without a trained vehicle operator--require a high level of scrutiny, and each assessment is tailored to the risks that are relevant to the intended operating mode. Of course, the level of detail in which we explore the capabilities of the AV is the highest when removing the trained operator due to the absence of the operator as a risk mitigation.

Waymo's process for making these readiness determinations entails an ordered examination of the relevant outputs from all of our safety methodologies combined with careful safety and engineering judgment focused on the specific facts relevant for a particular determination. Waymo will approve when it determines the ADS is ready for the new conditions without creating any unreasonable risks to safety. Waymo will continue to apply and adapt those methodologies, and to learn from the important contributions of others in the AV industry, as we continue to build an ever safer and more able ADS.

# Introduction

Waymo is the world's most experienced developer of automated driving systems ("ADSs"),[3] which, when installed in a vehicle, can perform the entire dynamic driving task ("DDT")[4] without human intervention. Waymo's safety philosophy is clear: *Waymo will reduce traffic injuries and fatalities by driving safely and responsibly, and will carefully manage risk as we scale our operations*. This philosophy provides the foundation for all of our activities, is consistent with having safety at the center of Waymo's corporate culture, and sets the course for Waymo's future.

---

[3] "AUTOMATED DRIVING SYSTEM (ADS). The hardware and software that are collectively capable of performing the entire DDT [Dynamic Driving Task] on a sustained basis, regardless of whether it is limited to a specific operational design domain (ODD); this term is used specifically to describe a level 3, 4, or 5 driving automation system." J3016 *Taxonomy and Definitions for Terms Related to Driving Automation Systems for On-Road Motor Vehicles* (June 2018) ("J3016") at 3.2.

[4] "DYNAMIC DRIVING TASK (DDT). All of the real-time operational and tactical functions required to operate a vehicle in on-road traffic, excluding the strategic functions such as trip scheduling and selection of destinations and waypoints, and including without limitation: Lateral vehicle motion control via steering

In order for automated vehicles ("AVs", which is used here to refer to any vehicle equipped with an ADS) to gain public acceptance and fulfill their promise for safety and mobility, the public must have a better understanding of their safety. This paper summarizes how Waymo builds safety into our AVs and determines their safety readiness for deployment. Waymo's 2017 Safety Report was the first public safety report issued by an AV developer and provided a description of our technology and safety processes at that time. We have recently issued an updated version of that report.[5] In order to promote transparency and offer the benefit of Waymo's experience, this summary provides a closer look at our current safety processes and methodologies and how they relate to each other. Waymo seeks to improve these methodologies and processes continuously as we continue to learn.

Waymo produces SAE[6] Level 4 ADSs[7], which makes Waymo's vehicles equipped with those ADSs capable of driverless operation. Waymo began as the Google Self-driving Car Project in 2009, and our early testing of automated driving systems that relied on human intervention for safety convinced us that Level 4 automation--which does not rely on a human driver for any reason--was the best option for achieving enhanced safety through automation.

Waymo's safety methodologies, which draw on well established engineering processes and address new safety challenges specific to AV technology, provide a firm foundation for safe deployment of our Level 4 ADS, which we also refer to as the Waymo Driver. Waymo's determination of its readiness to deploy its AVs safely in different settings rests on that firm foundation and on a thorough analysis of risks specific to a particular Operational Design Domain ("ODD").[8] Understanding the ODD concept is essential to understanding ADS safety.

---

(operational); Longitudinal vehicle motion control via acceleration and deceleration (operational); Monitoring the driving environment via object and event detection, recognition, classification, and response preparation (operational and tactical); Object and event response execution (operational and tactical); Maneuver planning (tactical); and Enhancing conspicuity via lighting, signaling and gesturing, etc. (tactical)." J3016 at 3.13.

[5] Waymo Safety Report (September 2020).

[6] SAE International, which began as the Society of Automotive Engineers, is a global association of more than 128,000 engineers and related technical experts in the aerospace, automotive and commercial-vehicle industries. Its programs include education on technical subjects and voluntary consensus standards development.

[7] "LEVEL or CATEGORY 4 - HIGH DRIVING AUTOMATION. The sustained and ODD-specific performance by an ADS of the entire DDT and DDT fallback, without any expectation that a user will respond to a request to intervene." J3016 at 5.5.

[8] "OPERATIONAL DESIGN DOMAIN (ODD). Operating conditions under which a given driving automation system or feature thereof is specifically designed to function, including, but not limited to,

An ADS (except at SAE Level 5) is specifically designed to perform the entire DDT only in specific operating conditions related to environmental, geographic, roadway and other conditions. Accordingly, development, analysis and measurement of ADS safety must address the particular ODD of the given ADS.

This paper is being released in parallel with a paper on Waymo safety performance data relevant to the driving experience of our ADS thus far.

# Waymo's Current Automated Vehicle Operations

Waymo currently operates our AVs primarily in California and Arizona, with additional testing in several other states including Michigan,Texas, Florida and Washington. In Metro Phoenix, our fleet of Chrysler Pacifica AVs has been transporting passengers in various types of self-driving service since 2017, including a driverless service. Waymo uses "driverless" here to refer to operations in which the ADS controls the vehicle for the entire trip without a human driver (whether in the vehicle or at a remote location) expected to assume any part of the driving task.[9] We expand operations to new locations and new operating conditions gradually, starting with manual driving, then automated driving with an onboard trained vehicle operator who can take control. We move to driverless operation only when we have carefully determined the readiness of our AV for safe driverless operation in those new situations. "Self-driving" miles, as Waymo uses the term, include both driverless operation and miles in which the ADS controls the vehicle but a trained vehicle operator is present with the ability to assume the driving task.

Through September 2020, Waymo has compiled over 20 million self-driving miles on public roads operating in over 25 cities, including 74,000 miles in driverless operation. As discussed below, this on-road experience is one important part of our overall safety process.

# Approaches to Assessing Automated Vehicle Safety

Some recent surveys suggest that the public has concerns about the safety of AVs, and the question of "how safe is safe enough for AVs" has received a good deal of attention. Various public and private organizations have recently offered recommendations on how the safety of AVs might be measured and demonstrated, and they have suggested the need for a common vocabulary on AV safety to promote better understanding of the subject. This section briefly

---

environmental, geographical, and time-of-day restrictions, and/or the requisite presence or absence of certain traffic or roadway characteristics." J3016 at 3.22.

[9] This definition is in accord with the SAE J3016 definition: "3.11 DRIVERLESS OPERATION [OF AN ADS-EQUIPPED VEHICLE]. Operation of an ADS-equipped vehicle in which either no on-board user is present, or in which on-board users are not drivers or fallback-ready users." The definition used in this paper makes clear that driverless operation does not include remote control operation by a human. Waymo does not use remote control (teleoperation) to operate its AVs. Our Fleet Response team can provide information and direction to the ADS, which still performs the entire DDT.

reviews some of the important efforts to develop approaches that might improve public understanding of AVs and increase public confidence in AV safety. Although Waymo does not find any single approach fully sufficient, many are worthy of consideration for the insights they convey.

Billions of people have traveled trillions of miles in motor vehicles in the last 120 years. In nearly every case a human driver has controlled the vehicle. Given the weight of this human experience, it is not surprising that the public and regulators are interested in understanding how companies that are developing driverless vehicles are working to address their safety and on what basis they are making safety determinations. The thousands of people who have already had the opportunity to ride in Waymo's vehicles in self-driving mode--including those who have used those vehicles repeatedly in Waymo's passenger service in Arizona--have directly experienced the safety of those vehicles.[10] Our riders and those who have not yet had the opportunity to experience AVs firsthand may reasonably desire a better understanding of the processes used to help ensure the safety of driverless operation. The Waymo Safety Report and this paper on our safety methodologies and readiness determinations aim to contribute to that better understanding.

This understandable desire for more information about how developers validate and demonstrate AV safety has led to many attempts to articulate safety processes, principles and measurement methods for AVs. Various private and governmental organizations have proposed a wide range of methodologies for measuring or demonstrating AV safety. For example, in 2018, the National Highway Traffic Safety Administration ("NHTSA") published a frequently cited research report[11] explaining in some detail how performance tests for ADS safety can be developed in ways that are focused on the specific ODD of the ADS. The report included illustrations of how tests could be performed on several specific types of ADS features (e.g., a Level 4 passenger transportation service).[12] The report provides a great deal of important information on relevant concepts and illustrates the complexities of demonstrating ADS safety in an ODD-specific manner and the important role of simulation in determining ADS safety.

Some organizations have provided helpful documents that attempt to define important terms related to AVs, articulate basic safety principles for ADSs, or flesh out the meaning of basic concepts in order to promote a deeper understanding of how those concepts may be applied in demonstrating ADS safety. The leading example is the SAE's Recommended Practice J3016, Taxonomy and Definitions on automated driving.[13] This seminal document sets out recommendations for defining the various levels of automation and related terms. The SAE's

---

[10] For example, by early 2020, we were providing between 1,000 to 2,000 rides per week in our Waymo One service in Arizona, with 5 to 10 percent of the rides in any given week driverless.
[11] *A Framework for Automated Driving System Testable Cases and Scenarios* (September 2018).
[12] Although NHTSA has not yet proposed new standards for ADS safety, the United Nations Economic Commission for Europe has begun a process to develop such standards. See *Revised Framework Document on Automated/Autonomous Vehicles*, ECE/TRANS/WP.29/2019/34/Rev.2 (December 2019).
[13] J3016 *Taxonomy and Definitions for Terms Related to Driving Automation Systems for On-Road Motor Vehicles* (June 2018).

Automated Vehicle Safety Consortium has issued various documents elaborating on certain AV safety concepts, including a best practice for developing detailed descriptions of the various elements of an ODD.[14] These efforts and others seek to provide the conceptual and terminological clarity that is difficult to achieve but is very helpful if ADS safety is to be demonstrated effectively.

Underwriters Laboratories ("UL") and Edge Case Research have produced *UL 4600*,[15] which is a suggested standard for evaluation of autonomous products, specifically vehicles at SAE Levels 3 through 5 (i.e., those equipped with an ADS). The standard describes a "safety case approach" for demonstrating AV safety that is goal-based and technology-agnostic. The safety case articulates specific goals and evidence-based arguments on how the ADS meets those goals. The overall objective is to demonstrate via documentation that an ADS is "acceptably safe," a recognition that transportation always entails risk and that realistic goals involve *avoidance of unreasonable risks*, not all risk. Although UL 4600 suggests how some aspects of traditional engineering and system safety methodologies can be useful, the standard is explicitly non-prescriptive with regard to any overall design process or safety methodology.

Other organizations have recommended AV safety methodologies that employ formal mathematical models to define acceptable performance of an ADS with regard to maintaining a safe distance from other road users within a spatial envelope. Under these approaches, ADS safety is demonstrated by showing the ADS's ability to adhere to certain mathematical rules that ensure an acceptable level of risk in particular scenarios. One challenge with these approaches is developing a consensus on the appropriate values to use as inputs to the mathematical rules in a wide range of traffic scenarios, a problem that is magnified as ODDs scale to include more complex scenarios. Also, some of these "envelope" methodologies do not wholly address the ADS's ability to take reasonable actions, if circumstances permit, in response to unreasonable actions of other road users, focusing instead on how the ADS must perform to avoid being considered at fault.

A group of vehicle manufacturers and electronic systems suppliers have issued a thorough summary of AV design and validation methodologies called *Safety First for Automated Driving*.[16] This report notes the challenges in demonstrating the safety of Level 3 and 4 ADSs (e.g., the impossibility of testing for every conceivable driving scenario the ADS may encounter) and proposes a thoughtful approach to addressing those challenges, including extensive use of various types of simulation in system and subsystem validation.

The RAND Corporation produced an extensive report recommending a general framework for measuring AV safety.[17] The report notes that measures such as actual crashes are dependent on exposure over large accumulations of mileage, which is rare for AVs at this stage of their development. The report makes clear how challenging it can be to make statistically valid comparisons between the performance of AVs and human-driven vehicles, particularly in identifying comparable data that is ODD-specific. The report recommends some methodologies that can be used to measure AV safety during development, demonstration, and deployment. A different RAND study[18] contends that delaying full deployment of AVs until an extraordinarily high level of safety is achieved in comparison to human drivers could cost hundreds of thousands of lives over many years.

---

[14] *AVSC Best Practice for Describing an Operational Design Domain: Conceptual Framework and Lexicon* (April 2020).
[15] UL 4600 Standard for Evaluation of Autonomous Products
[16] *Safety First for Automated Driving* (2019).
[17] *Measuring Automated Vehicle Safety, Forging a Framework* (2018).

One frequent subject in the relevant literature is how to measure ADS safety in comparison to human driver safety. No consensus has emerged on that point. Most writers underscore the significant challenges in identifying the appropriate data on human driving behavior that matches closely to the specific ODD for which an ADS is designed. Also, uncertainty exists in selecting data that provides a statistically sound basis for predicting the frequency and severity of accidents that may result from the AV's interactions with human drivers and other road users in a specific set of driving conditions. We explain later in this paper how Waymo attempts to address these challenges.

Our process is informed by our understanding of certain system engineering methodologies of special importance in the automotive industry.[19] For example, ISO Standard 26262[20] provides guidelines for identifying, categorizing, and addressing hazards caused by malfunctions in safety-related electrical or electronic systems in passenger vehicles over the life cycle of those systems. The goal of applying the standard is to avoid or mitigate the effects of system failures in order to ensure "functional safety," which the standard defines as the "absence of unreasonable risk" due to potential harm that may be caused by such failures. The standard categorizes the hazards by risk levels (automotive safety integrity levels, or ASILS) and includes an approach for developing and validating requirements to ensure that an acceptable safety level is achieved for each item being analyzed. The probability, severity, and controllability of each hazard must be considered in determining its risk level, which triggers the appropriate action to address the hazard. ISO 26262 has provided significant insights for Waymo's hazard analysis processes. However, Waymo does not rely strictly or exclusively on ISO 26262's principles, which are not a perfect fit for a Level 4 ADS, where there is a need for a special focus on the plethora of conditions likely to be encountered in the intended ODD, and where separate analysis of individual items may not be as useful as analysis of hazards related to system interactions.

---

[18] *The Enemy of Good, Estimating the Cost of Waiting for Nearly Perfect Automated Vehicles* (2017).
[19] An excellent comparative examination of leading hazard analysis and risk assessment methodologies as applied in the auto industry is found in *Assessment of Safety Standards for Automotive Electronic Control Systems* (Hommes, 2016), a report prepared for NHTSA by the Volpe National Transportation Systems Center.
[20] ISO 26262 -1 (2018), *Road Vehicles--Functional Safety.*

Another auto industry standard, ISO/PAS 21448,[21] focuses on the "safety of the intended functionality" ("SOTIF"), rather than on avoiding and mitigating the effects of system failures. In other words, it asks whether a system's intended behavior is safe in the likely conditions of its use. Similar to ISO 26262, SOTIF defines safety as the absence of unreasonable risk, but in this case the risk relates to hazards caused by the performance or limitations of the intended functionality. The standard offers guidelines for assessing hazards that may arise from insufficiencies in the system's intended functionality or from foreseeable misuse of the system. SOTIF is designed for analyzing emergency intervention systems and advanced driver assistance systems (SAE Levels 1 and 2) in vehicles, but the standard notes that it "can be considered for higher levels of automation, however additional measures might be necessary."[22] Accordingly, while it provides helpful insights into how even the intended functions of highly complex automotive systems can entail hazards that need to be identified and addressed, the SOTIF standard is not intended to be solely sufficient for ensuring the safety of higher levels of automation (such as Waymo's Level 4 ADS). Waymo's safety analyses are consistent with, but not dependent on, SOTIF principles.

Waymo is participating in activities that seek to build on and add to the existing methodologies and metrics for measuring AV safety and best practices for ensuring AV safety. Those efforts include participation in various SAE task forces as well as international regulatory initiatives focused on AV safety.

Several common themes emerge from these various approaches: (1) ADS safety can be measured effectively only with a careful focus on testing capabilities related to the specific ODD for which the ADS is designed; (2) simulation-based testing is an extremely important part of development and safety validation of an ADS; (3) meaningful comparisons between ADS safety and human driving are very challenging to derive from available data; (4) testing for every conceivable scenario is impossible, so scenario selection should be tailored for risks likely to occur in a given ODD; and (5) as in all transportation, transportation in an ADS-equipped vehicle involves some level of risk, and a guiding principle should be to avoid risks that are unreasonable.

As insightful and well conceived as many of these suggested and developing methodologies are, none of them provides a definitive, widely accepted, empirical methodology for answering the question often asked with regard to AVs: "How safe is safe enough?" Moreover, no consensus exists on a single metric or methodology to demonstrate that an AV is safe. Accordingly, Waymo continues to learn from these various approaches but relies on our own combination of methodologies to help ensure the ADS performs in a reasonably safe manner in its driving environment.

---

[21] ISO/PAS 21448: 2019, *Road Vehicles--Safety of the Intended Functionality.*
[22] Id. at "Scope."

# Waymo's Safety Methodologies Work Together to Reduce Risk

Waymo's safety methodologies focus on the development, qualification, deployment and sustained field operation of a unique product: a Level 4 ADS that can perform the entire DDT, with no human driver present, in both normal traffic conditions and very challenging scenarios that we have reason to expect may occur in a specific ODD. Therefore, while we learn from certain widely accepted engineering processes and principles (including those discussed above), we tailor them--as informed by our extensive experience in Level 4 technology--for this purpose. We continuously refine those methodologies in an incremental way as we scale our operations.

In addition to having completed over 20 million miles of public road testing[23], Waymo's AV experience includes billions of miles of driverless operation in detailed simulation, more than a decade of structured testing in test facilities, and constant analysis to identify and address serious safety hazards. That experience has taught us that no single safety methodology is sufficiently holistic; instead, multiple methodologies working in concert are needed. Accordingly, our safety framework includes a variety of complementary methodologies that Waymo seeks to improve continuously and that reflect our safety philosophy, which focuses on how we will reduce traffic injuries and fatalities and manage risk as we scale our operations.

Waymo's approach to safety focuses not just on the ADS but also considers the safety of the base vehicle and the safety of our AV operations. Waymo's methodologies address the full life cycle of our AVs, from design and development, through phases of testing, and during and after deployment.

Waymo's various safety methodologies are supported by a portfolio of test types. At the system level, we have three basic types of testing: simulation, closed-course, and public road. At the subsystem and component level, we conduct software unit tests, integration tests, bench tests and other hardware-in-the-loop tests. These types of testing are in constant interaction; each informs and complements the other. Closed-course testing, for example, uses test scenarios derived from multiple sources (e.g., real-world test driving on public roads, crash databases, and naturalistic driving studies) and is used to build and validate elements of our simulation models.

The miles we drive on public roads are extremely valuable for the advancement of our ADS. The lessons learned from our extensive on-road experience are critical and inform all aspects of our safety program. Not only does on-road testing provide new scenarios for structured, closed-course testing and help validate our simulation models, those miles can test basic ADS competencies, provide empirical testing of various metrics, and reveal previously unknown risks.

---

[23] In its report *Driving to Safety* (2016), RAND notes its estimate that it would take hundreds of millions or perhaps billions of miles of highway driving to demonstrate AV safety with driving miles alone and voices support for innovative methods, including simulation, for demonstrating AV safety.

Our advanced simulation capabilities are also a very important element of our safety methodologies. Simulation, of course, can test ADS capabilities involving thousands of scenarios and road user interactions that would take far longer to experience in on-road testing and be very difficult and perhaps unsafe to replicate in structured testing.

One valuable aspect of our testing programs is how our real world miles and our scaled simulation work together to improve each other. We are proud of the years of on-road experience that we have developed, and how that experience has informed our understanding of the difficulties of driving and also the promise of our technology. But on-road driving has its limitations, in particular the fact that it is relatively slow and expensive. Thus we have also developed a complementary simulator that can accelerate our development and testing inspired in part by our on-road experience, and expanded to represent even more situations than we see in real life. However, any simulation requires validation, so we rely on road driving to validate our simulator, thus completing a virtuous cycle of learning and refinement.

Waymo's safety process is a continuous one in which lessons and feedback from one element or stage are used to foster improvement (e.g., calibration or correlation) in other elements and stages. This includes review of any issues of concern revealed by any safety methodology or in any test mode, prioritization of their importance, and disposition of the highest priority items.

Driverless vehicles and their ADSs are incredibly complex undertakings, blending advanced hardware technology, cutting edge artificial intelligence and large fleet operations. Developing each of these elements requires leveraging and adapting different engineering practices and evaluation methods to ensure the best outcomes. Waymo thoughtfully applies different methods to different types of technology in different stages of the life-cycle, and while no hard boundaries are drawn, we attempt to describe the essence of those methods and techniques below under headings that refer to these three layers of our technology: hardware, ADS behavior, and vehicle operations. Some of our methodologies (e.g., hazard analysis) are used across the different layers of our technology. Therefore, rather than discuss the methodologies in isolation, we discuss how each is used in each of those technology layers. Figure 1 provides a visual representation of our technology layers, the critical attributes for each layer, and the primary methods we use to ensure their safety.

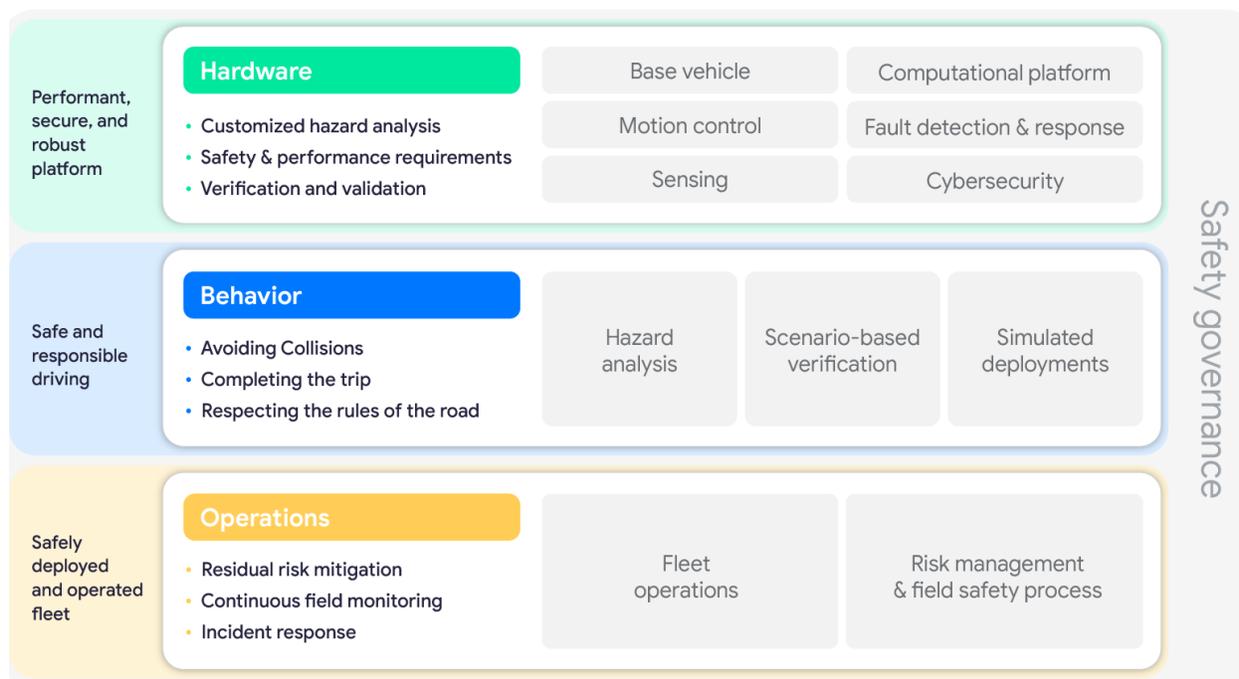

*Figure 1. Schematic representation of Waymo's three technology
layers and the methodologies presented in this paper*

## The Hardware Layer

The hardware layer refers to four major subsystems of the self driving vehicle: the base vehicle platform, the motion control actuators, the attached suite of sensors, and the computational platform used to run our advanced software. Each of these subsystems is important to both the safety and performance of the AV and ADS. Also relevant to the hardware layer (and in fact the whole system) are the steps we take to ensure the protection of our ADS from system or subsystem faults and cybersecurity threats.

To ensure the development of a safe and robust hardware layer, we primarily use a systems engineering approach that defines requirements for the performance and safety of the aforementioned sub-systems and also includes thorough verification and validation of the components, sub-systems and the fully integrated systems. This method, which is heavily adopted in both automotive and aerospace industries, is particularly effective in defining and testing deterministic behaviors and well-defined performance characteristics, which are common in the hardware layer. To execute this approach well, one must be thoughtful about the derivation of the requirements and thorough in the testing to ensure that the system is adequately specified, designed and tested to meet the desired task.

To thoroughly develop the requirements for the hardware layer, we focus on both performance and safety requirements. Safety requirements are often defined through various safety processes such as Design Failure Mode and Effects Analysis (DFMEA), Hazard Analysis, and Fault Tree Analysis. Waymo uses these processes to develop a set of requirements that can ensure the safe operation of the AV even in the presence of fault or error.

Independent of the method of derivation, the requirements established for the hardware layer are both verified and validated at multiple stages of the program. Verification and validation ("V&V") are often discussed as a single methodology, although the terms refer to distinct parts of an engineering process. In brief, verification refers to the process of determining that a system or component meets a specific requirement or specification, while validation refers to determining, usually through some form of testing, that the verified system or component serves its intended purpose. Verifying that the hardware elements satisfy requirements and validating that they serve their intended purpose play a crucial role in ensuring that our ADS has fundamental behavioral competencies to perform safely. We use various evaluation methods, including simulation-based and hardware-in-the-loop testing and analysis, structured testing with a fully-integrated vehicle, and the analysis of field data.

## Base Vehicle

The base vehicle platform establishes the foundation for the physical safety of the passengers. We begin by purchasing very safe, fully certified vehicles from experienced vehicle manufacturers.[24] We have purchased hundreds of Chrysler Pacifica Hybrid minivans from FCA. Those vehicles, which comprise nearly all of Waymo's current passenger fleet, have a 5-star safety rating (the best) from NHTSA as well as a "Top Safety Pick" rating from the Insurance Institute for Highway Safety.[25] These ratings primarily demonstrate the excellent crashworthiness of these vehicles. We are currently adding the all-electric Jaguar I-PACE to the Waymo fleet. Although not yet rated in the U.S., the most recent model year of the I-PACE that has been rated in Europe received a 5-star rating from the European New Car Assessment Program.[26] Our AVs are among the safest vehicles on the road as measured by these rating programs, particularly with regard to the vehicle's ability to protect occupants from injury in a crash. Moreover, we work to ensure that the modifications Waymo makes to the base vehicles (discussed below) do not reduce the crashworthiness of those vehicles. We also extend the capabilities and robustness of the base vehicle for Level 4 autonomous driving by upgrading the redundancy and capacity of the power, networking, and thermal systems, helping to ensure that the Waymo ADS and associated base vehicle systems are available for safe driving and that a minimal risk condition can be achieved.

---

[24] Waymo has manufactured vehicles itself. In 2014 and 2015, Waymo manufactured 61 low-speed AVs of a completely unique design. Waymo (then in the form of Google Auto LLC) certified these vehicles (except for some not used on public roads) as compliant with FMVSS No. 500 (Low-speed Vehicles). These vehicles, which could operate without the use of traditional manual controls, provided an outstanding platform for developing and testing driverless technology on closed courses and public roads. Having served that purpose, these vehicles are no longer in service.

[25] NHTSA's NCAP rating for the 2018 Pacifica is here. The IIHS rating for the 2018 Pacific is here.

[26] Euro NCAP's rating for the 2018 iPace is here.

## Motion Control

As part of development of our base vehicles with our OEM partners and automotive suppliers, we specify the inclusion of redundant braking and steering actuators, which we feel is necessary for safe, driverless vehicles given the failure rates seen in available non-redundant systems. In human-driven vehicles, the human driver is available to physically steer or brake the vehicle even in the presence of failures of the system or driver assist features. However, in driverless operation, the ADS must be able to detect a failure in a portion of the steering or braking system and still maintain sufficient control of the vehicle to bring the vehicle to a safe stop after that fault. The performance and fault tolerance of these motion control actuators is dictated by a thorough set of technical requirements, specifying performance in both nominal and faulted conditions, and backed by extensive verification including hardware-in-the-loop and closed course testing, as well as validated on closed courses and in the field.

## Sensing

Starting with the base vehicle and motion control systems, Waymo adds our own sensing systems, including a multitude of lidar, radar, camera, inertial and audio units that provide an expansive understanding of the driving environment and of the vehicle's precise location at any given point in time. Similar to the motion control system, these sensing systems are designed to meet rigorously defined performance and safety requirements. The performance requirements (e.g., the range of the forward sensing system) are typically developed through the decomposition of the intended use cases or driving maneuvers given the ODD. For example, an ODD that entails driving at highway speeds will decompose into a forward sensing range requirement that is greater than the range needed for driving on residential roads with a maximum speed of 25 m.p.h. While it is possible that a given hardware configuration meets both of those requirements, we actively verify our system against the requirements for the intended ODD.

These performance requirements are developed through simulation and analysis, but also leverage the 10 years and 20+ million miles of public roads driving that Waymo has experienced. The performance requirements on the sensing systems are particularly important to ensure that the decision-making layers of the software have adequate and accurate information on which to make driving decisions. The safety requirements, as derived through appropriate safety processes, ensure fault tolerance of the overall system, by detecting a failure of any given sensor and either safely ending the mission or having adequate redundancy to maintain sufficient performance. In addition to performance and safety requirements, we also have reliability and durability requirements that ensure the long term stability and performance of our fleet. Each of these sets of requirements is verified at appropriate stages of the program, using many methods including simulation, bench testing, and field testing.

## Computational Platform

To run our advanced behavioral software that is discussed in the next section, Waymo has developed a state-of-the-art computational platform that combines extreme performance with proven reliability and fault tolerance. The computational demands of advanced artificial intelligence are immense, and we design our computational platform to meet the needs of executing the World's Most Experienced Driver™, as we sometimes refer to our ADS. In addition to the advanced performance, we have developed this computer system to have the appropriate amount of redundancy. In case of the failure of the main computer, there is an additional computer system that has the ability to bring the vehicle to a safe stop. In addition, we have made and tested multiple design choices (like redundant power systems) to help ensure the availability of the computer system.

## Fault Detection and Response

In the discussion above of elements of the hardware layer, we mentioned briefly the relevant fault tolerance requirements and the associated verification. As discussed below, the behavioral layer also includes V&V strategies to address detection and response of software faults. The importance of this fault detection and response attribute of our system cannot be overstated. A Level 4 ADS, by definition, must be able to achieve a minimal risk condition[27] without human intervention in the event of a failure of either the ADS or another vehicle system that prevents the ADS from performing the dynamic driving task or upon encountering circumstances outside of the approved ODD's limitations.[28] In addition to the fault detection and fault tolerance mentioned in the sections above, we have designed our system to have a portfolio of appropriate fault responses given any failure. The proper response will vary according to the type and extent of the system failure and the immediately surrounding traffic situation.

The response to faults, accordingly, involves a flexible strategy that considers the relative risks associated with continuing operation, pulling off the road, or stopping in place. For example, in the case of a severe and total system failure, the lower level computers will engage the brakes and achieve the minimal risk condition. However in a less severe degradation of performance, the system can self-detect the degradation and the behavioral layer will still maintain control of the vehicle and actively drive to an appropriate location for further action (like parking on a side street for roadside assistance). The overall fault detection and fault response systems are specified and verified through the systems engineering process and undergo continuous validation to ensure that no requirements are missed.

---

[27] "MINIMAL RISK CONDITION. A condition to which a user or an ADS may bring a vehicle after performing the DDT fallback in order to reduce the risk of a crash when a given trip cannot or should not be completed." J3016 at 3.17.
[28] "DYNAMIC DRIVING TASK (DDT). All of the real-time operational and tactical functions required to operate a vehicle in on-road traffic, excluding the strategic functions such as trip scheduling and selection of destinations and waypoints, and including without limitation: Lateral vehicle motion control via steering (operational); Longitudinal vehicle motion control via acceleration and deceleration (operational); Monitoring the driving environment via object and event detection, recognition, classification, and response preparation (operational and tactical); Object and event response execution (operational and tactical); Maneuver planning (tactical); and Enhancing conspicuity via lighting, signaling and gesturing, etc. (tactical)." SAE J3016 at 3.13.

## Cybersecurity

While fault tolerance can be used to describe the robustness of a system against internal failures, cybersecurity involves the robustness of a system's avoidance and detection of and response to external attack. The modern world has shown us that attractive targets are frequently attacked, and sound security engineering practices are required to create a system that is robust and reasonably reduces cybersecurity risks.

Waymo has developed such a robust process to identify, prioritize, and mitigate cybersecurity threats. Although addressed here in the hardware section, our process extends across the software and operations layers as well. As an Alphabet company, Waymo's security practices are built on the foundation of Google's Security processes and are informed by publications like the NHTSA cybersecurity guidance[29] and the Automotive Information Sharing and Analysis Center's (Auto-ISAC) *Automotive Cybersecurity Best Practices*.[30] To help develop future security best practices, Waymo has also joined the Auto-ISAC, an industry-operated initiative created to enhance cybersecurity awareness and collaboration across the global automotive industry.

Our security review process examines threats ranging from fleet-wide remote attacks to single vehicle physical attacks as they relate to a driverless transportation service. A risk-based approach is used to prioritize mitigations. Our current process entails a comprehensive review of potential security access points to our ADS from both the interior and exterior of the physical vehicle, and we take steps to limit the number and function of those access points.

This begins by collaborating with our OEM partners at the onset to identify and mitigate vulnerabilities of the base vehicle. In addition, our continually improving software and vehicle design processes incorporate cybersecurity risk assessments, allowing us to implement defenses and protections according to the risk posed by each known vulnerability. New software releases go through an extensive review and verification process. Our hazard analysis and risk assessment processes have been designed to identify and mitigate risks that might affect safety, including modifications now being implemented to ensure coverage of risks related to cybersecurity.

We use layers of security to protect our ADS, especially safety-critical functions like steering and braking, against unauthorized communications, including vehicle control commands. We also consider the security of our wireless communications. Our vehicles do not rely on a constant communications connection to operate safely. While on the road, all communications (e.g., redundant cellular connections) between the vehicles and Waymo are encrypted, including those between Waymo's operations support staff and our riders. Our vehicles can communicate with our operations team to gather more information about road conditions, while our ADS maintains responsibility for the driving task at all times.

---

[29] Cybersecurity Best Practices for Modern Vehicles.
[30] Executive Summary of Auto-ISAC's Automotive Cybersecurity Best Practices.

These protections help safeguard against remote attacks and threats from passengers or malicious actors in close proximity. Should we become aware, whether through our ADS or other means, of an indication that someone has attempted to impair our vehicle's security, Waymo will trigger its cross-functional incident response procedure, which involves impact assessment, containment, recovery, and remediation.

## The ADS Behavioral Layer

The behavioral layer describes the software that is capable of directing safe driverless movement of our AVs on public roads. Having started as the Google Self-Driving Car Project, our software development relies to some degree on Google's world-class software development practices, adapted in our case for application to safety-relevant technology.

Unlike the hardware layer, where more traditional, deterministic safety methodologies can readily be used to measure safety, the behavioral layer requires solving for an infinitely variable set of inputs (e.g., actions of other road users and roadway conditions). The complexity of the challenge requires the use of sophisticated algorithms and specialized evaluation methodologies to determine how well those algorithms perform. There are three primary capabilities on which we evaluate the performance of the ADS's behavioral layer: avoidance of crashes, completion of trips in driverless mode, and adherence to applicable driving rules. In each case, we use the various methodologies described below to evaluate those capabilities in the context of the ODD in which the ADS is designed to operate.

To fully develop and understand the performance of the behavioral layer, we use a multi-faceted approach. Our approach begins with hazard analysis, by which robustness is built into our designs from the beginning. We then heavily leverage scenario-based verification, to ensure that the ADS behavior is in line with our requirements and expectations. Finally, we subject our system to large scale simulated deployments which allow us to empirically measure aggregate performance metrics.

### Hazard Analysis

Waymo leverages hazard analysis techniques to develop and test safety-critical software in line with established industry best-practices. Hazard analysis is a well-established methodology used to identify potential causes of safety risks and either eliminate or mitigate those hazards early in the engineering process. In the past decades, reliability and safety professionals have developed many hazard analysis techniques. These include deductive methods such as Fault Tree Analysis (FTA) and System-theoretic Process Analysis (STPA), and inductive methods such as Failure Mode and Effects Analysis (FMEA). Depending on the systems being analyzed, such as hardware, behavioral software or embedded controls software, and the development stage that these systems are in, Waymo chooses the most effective and efficient method to provide early identification of potential causes of safety issues, identify mitigations, prioritize the mitigations, develop safety requirements for the mitigations, and verify that the mitigations meet the requirements.

For example, we found STPA[31] is important for analyzing hazards in complex systems. Although not designed specifically for automotive systems or ADSs, STPA provides a method to identify, analyze, and mitigate hazards that may arise from non-obvious component interactions and lead to accidents. STPA, which is based on system theory, may be better able than more traditional hazard analysis methods to spot dangerous interactions that could create risks, whether due to a functional failure or a deficiency in the intended functionality.

Sometimes, existing system safety methodologies are not sufficient and we develop customized approaches to better fit our own needs. For example, for behavior software subsystems, Waymo developed a customized software subsystem hazard analysis method that is an integral part of the driverless software design and release process.

## Scenario-based Verification Programs

Waymo uses a variety of scenario-based testing approaches to ensure the ADS is capable of basic behavioral competencies as well as certain advanced functionalities. The competencies and functionalities to be tested in these test programs are derived from systematic methods of describing competencies needed in the ODD, naturalistic driving research data, and public road testing (real things that we observed in the field). The virtual test scenarios are either harvested from public road driving logs, obtained from closed course driving logs, or created from scratch in simulation-only workspaces. In addition, we also selectively conduct physical tests on candidate software on a closed course. These scenario-based test sets are used to evaluate the ADS's performance across the broad spectrum of conditions likely to be encountered in the ODD.

One example of such a testing program is our Collision Avoidance Testing Program. In addition to demonstrating an AV's capabilities in "normal" driving situations and in system failure conditions, an ADS, within reason, should have some level of ability to avoid or mitigate crashes in urgent situations relevant to the ODD and caused by the behavior of other road users. NHTSA, for example, recommends such testing.[32]

In addition to using many structured testing scenarios to examine the ADS's crash avoidance capabilities, our Collision Avoidance Testing Program also uses simulation to test ADS capabilities in thousands of scenarios where immediate braking and/or steering is required to avoid a collision. These scenarios include many involving vulnerable road users (e.g., pedestrians and cyclists) as well as other road users. Waymo seeded this test program with scenarios inspired by naturalistic driving data and crash databases.[33] Unlike the ADS's behavioral competencies in routine driving situations, these scenarios test competencies that may rarely need to be exercised but which are crucial in reducing the likelihood or severity of collisions induced by the behavior of other road users.

---

[31] *STPA Handbook* (2018).

[32] *Automated Driving Systems 2.0: A Vision for Safety* at 9. Similarly, the Insurance Institute for Highway Safety has developed specific tests for crash avoidance features such as automatic emergency braking and pedestrian emergency braking, and Euro NCAP is developing such tests.

Some of these scenarios may be sufficiently common to be considered part of the ADS's core competencies, while others may be considered edge cases (i.e., where one operating parameter is at an extreme value) or even corner cases (i.e., where more than one parameter is beyond normal conditions). What they have in common is their relevance to the intended ODD and the ability of the ADS to respond immediately to the actions of other road users to avoid or mitigate a crash.

Waymo assesses the ADS's performance in these scenarios against that of a simulated reference agent whose predicted performance (e.g., response time and steering/braking capability) is based on an analysis of naturalistic driving data of human driver performance in real-life situations. The rigor with which we devise individual scenarios and clusters of similar scenarios, and the extensive effort we employ to estimate the reference agent's performance, provide a sound comparison point for evaluating the performance of our ADS in similar scenarios. While the specific scenarios and metrics are proprietary, we believe that the representativeness of our scenarios and our performance criteria contribute significantly to the overall safety of our system.

Collision Avoidance Testing is just one example of a scenario-based testing program, but shows how we use the best available research, our on-road experience and systematic methods to craft testing programs that can evaluate critical aspects of our ADS's safety-relevant behavior. As with all of our methodologies, each of our scenario-based testing programs will continue to evolve as the available technology and research progresses.

## Simulated Deployments

While the structured development and test processes described above can effectively establish the evidence for a performant system, it is important to conduct additional steps to both validate the methods used to this point (e.g., did we select the correct scenarios?) as well as objectively measure the ADS against critical performance indicators. To achieve these goals, we consistently put our system through simulated deployments, in which we try to answer the question: if this system were deployed in driverless mode in this specific ODD, how would it perform?

---

[33] In addition to these sources of real-world scenarios, Waymo closely monitors developments in crash avoidance test scenarios and metrics that safety rating organizations such as the Insurance Institute for Highway Safety and the various New Car Assessment Programs (NCAPs) are using or considering. Although those scenarios and tests are currently focused on driver assistance systems or lower levels of automation than Waymo's Level 4 ADS, we find their scenario definitions and test procedures germane for evaluating collision avoidance tests for our AVs, particularly with regard to avoidance of collisions with vulnerable road users.

These simulated deployments can take place in two main methodological variants. The first variant is one that is executed by re-simulating a certain software version and system configuration against historical logs. This method has the advantage of being highly scalable, allowing for extreme acceleration of results, and gives the ability for individual developers to execute the evaluation themselves. To ensure the credibility of our simulations, we have an ongoing simulation credibility process.[34]

The second variant involves driving on public roads with the added supervision of a vehicle operator, and only simulating the points in time where the operator took control of the vehicle. This variant we refer to as public roads driving with counterfactual simulations. This approach has the advantage of having very high credibility, since the software version under evaluation was controlling the vehicle in real time for the vast majority of the time. This variant as applied to predicting potential collisions is discussed in depth in the Waymo white paper entitled, *Waymo Public Road Safety Performance Data*.[35] Waymo does not believe an evaluation based entirely on simulation or entirely on actual driving (whether through structured testing or through public roads driving) can provide sufficient information to make a confident readiness determination. By leveraging both simulation and real driving, we can achieve a level of scale not possible in real driving alone as well as a level of credibility not possible in simulation alone. Across both of these methods, we study carefully three critical parameters: avoidance of crashes, completion of driverless trips, and adherence to rules of the road.

With regard to safety, the most important attribute of the ADS is its ability to minimize the frequency and severity of collisions. Through our simulated deployments, we measure empirically the AV's estimated collision rates that would have occurred in that simulated deployment, to allow comparison of those rates against human driver benchmarks and other performance criteria. To measure the SDC performance and establish a human benchmark to enable this comparison, there are two main areas of research and applications of human behavior that are needed. The first is correctly understanding how humans would have reacted to our AV in any given simulation, which is critical to understanding if there would or would not have been a collision. This human reference model can take multiple forms, either a single point model or a probabilistic model involving a range of human behavioral responses, and Waymo employs both. The second area of research is to derive the human-driving benchmarks for a comparison of performance. A fair degree of uncertainty exists with regard to both of these human attributes.

---

[34] Waymo uses closed course testing to ensure that various assumptions used in our simulation model are in fact accurate representations of our AV's performance. For example, for our simulation to reflect how our AV would perform in a particular scenario requiring hard braking, we need to know that the simulation replicates the actual performance of our AV using the same braking profile. Here again, our methodologies intertwine rather than stand alone. Simulation is an important aspect of many of our methodologies, but its accuracy depends on effective V&V of actual system capabilities.

[35] Schwall, M., Daniel, T., Victor, T.W., Favaro, F, and Hohnhold, H. (2020). Waymo Public Road Safety Performance Data. www.waymo.com/safety.

For the first factor of the analysis, while modeling the predicted behavior of a nominal human driver in specific situations presents challenges, the naturalistic driving data and certain focused research (e.g., on typical driver braking reaction and execution times) provide helpful sources. To the extent that the actions of other types of road users (such as pedestrians and cyclists) must be estimated in certain scenarios, additional uncertainties are introduced. Nevertheless, Waymo has studied the available literature on the subject, as well as conducted its own research in order to develop a model of nominal human behavior that we use to infer the behavior of other agents on the road. This is an active area of research in Waymo and in the industry, and we anticipate that our understanding of the nuances of human behavior will continue to evolve.

For the second factor (i.e., determining a human driving benchmark), while many assert that AVs must be "at least as safe as a human driver," determining with any degree of certainty exactly what that goal means and how to measure it can be very difficult. To date, no one has presented a fully satisfactory set of metrics for making such a comparison to human driving across the range of ADS capabilities, use cases, and ODDs. Waymo has over the years grappled with the limitations of several of the proposed approaches (which are briefly summarized below) and developed solutions, informed by our own experience, to the problem of identifying particularized human driving benchmarks useful for assessing risk in specific ODDs.

Several options exist for determining the benchmark of human driving for purposes of comparison. The ADS's ability to pass a basic driving test similar to the one a new driver must pass to obtain a license would be instructive, but would constitute too low a bar to make a determination of the ADS's safety. The on-road portion of such a test assesses just basic driving skills in normal driving, not in crash-imminent situations or those involving vehicle system failures in which an ADS is expected to perform safely. Human drivers, of course, continue to learn after getting their first license and encounter more difficult driving scenarios at a very uneven rate.

Similarly, the ADS's inherent ability to avoid the types of human errors (e.g., intoxication, speeding, distraction, drowsiness, etc.) that lead to the vast majority of fatal crashes is a given but is not, standing alone, a sufficient measure of ADS safety. The likelihood that broad deployment of AVs will prevent tens of thousands of deaths that would otherwise occur annually just in the United States is a major factor warranting rapid development and deployment of AVs.[36] However, measuring the safety of a specific ADS cannot be limited to measuring ADS behavior against only the behavior of drivers at their worst. Were such an *a priori* measure the only test, ADS safety would not be in question.

---

[36] RAND, *The Enemy of Good, Estimating the Cost of Waiting for Nearly Perfect Automated Vehicles* (2017).

Therefore, Waymo seeks to determine a human benchmark reflective of a nominal safe driver's behavior in certain situations. Determining the human benchmark is not a simple matter given the wide range of driving capabilities just within the United States, not to mention other countries. Consider just the drivers within any person's circle of family and friends--it's likely that no two of those drivers have identical driving skills, experience or accident history. Available driving data provide a ready source of information about driving abilities and patterns. Although crash data and naturalistic driving data are abundant, at least in the United States, breaking the data down by the conditions that closely reflect those of the intended ODD is not simple.

Moreover, because meaningful comparisons can be made only by considering the severity of crash outcomes, vagaries in how severity is recorded in different crash data sets presents analytical challenges, as does the limited amount of data on the most severe events. Nevertheless, Waymo has utilized the available data, from a variety of sources, to establish human performance benchmarks that we can tailor for various ODDs.[37]

With a model of the likely human behavior in reaction to our AV's behavior and the simulated or known behavior of the ADS, Waymo can predict the ADS's involvement in collisions across various severity levels of possible outcomes in that ODD. To achieve this prediction, we start with the logs from our original public driving, and either simulate the entire run segment or simulate only the time after the operator took control of the vehicle (as discussed above). In either case, we model what each of the other agents would have done had the simulated Waymo ADS been driving in the simulated location at that time. In the case of a simulated collision, we evaluate the severity and other attributes of the collision. With this method, we can establish the performance of the Waymo ADS across the evaluated ODD.

Given the assessment of the ADS performance and the established human benchmark for collision rates, we have the ability to make comparisons at different severity levels and across different ODDs. Our analysis gives greatest weight to events involving the highest severity, including those that would involve injury to vulnerable road users, and we set benchmarks accordingly. Although our analysis involves a degree of uncertainty in light of the factors already highlighted, we feel this approach can provide a rich understanding of the safety and safety risks of deploying the ADS in a target ODD. Of course, these comparisons to human driving are just one among many of Waymo's safety methodologies, and we continuously work to refine and improve such comparisons. These criteria and the analysis undertaken to develop them are important for understanding possible safety risks in a given ODD, and tracking the ADS's performance against those criteria provides a valuable method of measuring the safety of that performance.

---

[37] Those who study this issue can differ in their assessment of any particular model's validity, which is why it will take time to develop a consensus model of human driving. In the meantime, Waymo's modeling reflects extensive experience and resources invested.

In addition to a deep study of the predicted collision rates in any given deployment, Waymo also takes care to ensure that our ADS respects the rules of the road. Waymo's evaluation of the ADS's compliance with road rules, similar to our evaluation of the ADS's likely crash avoidance capability, relies on scenario-based verification and validation through large scale evaluation of on-road and simulated performance. This combination of methods allows us to ensure test coverage across the broad range of ODD-relevant road rules (including rules that are not commonly encountered in real world driving).

The third behavior that Waymo and our passengers are deeply interested in is reaching our passengers' destination without delay. A delay could be caused either by an internal fault in the ADS, or by an unforeseen aspect of the environment. In the case of an internal fault, a Level 4 ADS must be able to reach a minimal risk condition under its own power and control. These capabilities were described in the fault protection section above. There are also certain unexpected cases where the ADS may not have the ability to proceed to the destination due to the occurrence of conditions outside the operational limitations of its ODD. For example, an ADS may get caught in an environmental condition that it was not designed for, at which point it should not proceed. As with system faults, our ADS has the capability to perceive such conditions and respond appropriately by achieving a minimal risk condition[38]. Regardless of the cause, it is important to track the rate of occurrence of these incomplete trips, as well as the cause of the issue and the details of the outcome. Similar to our analysis of possible collision rates in the ODD, we also utilize different design analysis methods and scenario testing in advance of the simulated deployments. Also similar to that collision-related analysis, this attribute is measured in simulation as well as on road, and each method helps to strengthen the other.

Throughout these simulated deployments (and also the scenario testing programs), we conduct millions of miles of simulated driverless operation every day, which sometimes provides new findings potentially related to safety (but unrelated to the original intent of the simulation). Our team assesses and dispositions all identified issues to understand the potential severity and, in the event that a potential high severity issue is identified, conducts further analysis and software work as warranted. Waymo uses conservative estimates of severity in analyzing these issues to ensure we give proper attention to risks of the highest potential severity even if the incident prompting consideration was not itself very severe. This method provides yet another layer of discovery and evaluation to reduce safety risk before we deploy in driverless.

---

[38] The ODDs of our current Level 4 ADSs include geographic as well as other limitations. Because the ADS will not accept an origin, destination, or routing outside of that geographic limitation, the ADS is operable only within the geographic limits of its ODD.

# The Operations Layer

## Fleet Operations

Application of Waymo's safety methodologies at the hardware and behavioral layers yields highly capable AVs. Operating those AVs in a fleet on public roads requires additional methodologies to manage risks that may arise during those operations as well as to derive data that can feed back into the hardware and behavioral levels to assure continuous improvement. In many ways, these final processes serve as the final layer of validation, ensuring that the product as built and the requirements as written meet the needs of the customers.

For example, in addition to the methodologies already discussed in this summary, which relate primarily to engineering issues affecting the safety of Waymo's ADS, Waymo's safety program ensures application of industry-leading safety practices in the operation of our AVs, such as a fatigue management program for our trained vehicle operators and coordination with law enforcement and emergency responders on how to deal safely with driverless vehicles. We plan and prepare extensively to ensure our incident response capabilities are ready to address a range of events. Our passengers have immediate voice access to our support team for assistance with any questions or problems they may have. Waymo also recognizes the importance of seat belt use in any vehicle and we take a number of steps to encourage our AV passengers to use their belts.

While Waymo's ADS is responsible for making every driving decision on the road, we have a fleet response team that can provide remote assistance to the ADS if needed. Our ADS is designed to recognize unexpected situations and contact our fleet response team, who can confirm what the ADS is seeing and provide additional contextual information. Our ADS only asks these questions to gain a deeper understanding of what it has already detected and perceived. In some situations our remote assistant may direct the AV to pull over and stop (e.g., due to a need to await further direction as a result of a road closure far ahead), but the ADS continues to perform the entire DDT. The remote assistant cannot remotely control the vehicle (i.e., Waymo does not use teleoperation to control its AVs). The ADS does not require remote assistance for anything that is safety-critical or latency-sensitive.

These operational capabilities, including incident response and fleet response, are designed with processes and rigor similar to those applied to elements of our hardware and software. As an example, we use STPA to systematically identify potential safety and operations risks and proactively mitigate them before operating on public roads. We document requirements and conduct various types of testing on these operational capabilities.

## Risk Management and the Field Safety Process

Despite the thoroughness of Waymo's safety methodologies discussed so far, a complete safety program must address the residual risks that may remain from those processes and new risks revealed in actual operations. Waymo's safety program encompasses the full life cycle of our AVs. Our attention to safety continues after deployment of a new platform or software release and includes a focused program for assessing and promptly addressing potential safety issues that may be revealed during operations.

Waymo's Risk Management Program ("RMP") identifies, prioritizes, and drives the resolution of potential safety issues before new or updated features or software are used on public roads or tested at our structured test facility. Issues are prioritized using a consistent framework for categorizing the residual risk and escalated according to the category, ensuring issues with the highest risk receive the most attention and appropriate level of review. Also, the RMP instructs the teams responsible for the specific matter to develop a path to mitigate and close each risk according to priority, promoting accountability for the resolution of these issues. For potential safety issues with higher levels of residual risk, Waymo treats such risks as urgent and takes prompt and appropriate action. Issues with less residual risk may warrant more gradual disposition on a specific timetable.

Our Field Safety Program complements the RMP by identifying and effectuating appropriate disposition of potential safety issues based on information collected after updated features or software have been released for driving on public roads. The field safety process collects and helps resolve potential safety concerns from many other sources, including employees, our riders, the public and suppliers. To encourage openness about safety issues, which is a central tenet of Waymo's safety culture, the Field Safety Program has multiple processes for submission of potential safety issues by anyone at Waymo. The program includes review of urgent issues by 24-7 on-call engineers who can restrict all or part of the fleet from operating. A cross-functional committee made up of representatives from our Safety, Engineering, Product, Operations and Legal teams reviews incoming information, directs the information to appropriate staff for analysis, and determines based on evaluation whether specific action is necessary. For example, the committee will consider the need for responsive actions upon learning that a manufacturer of the base vehicles to which Waymo has added its ADS has announced a safety recall or other action that could affect the safety of our AVs. The committee would analyze the relevant information and decide whether any operating restrictions would be appropriate while awaiting a permanent remedy from the manufacturer.

# Safety Governance

An effective safety program must include a governance process for making important safety decisions based on the outputs from the various safety methodologies. Waymo's governance process includes a tiered system of analyzing safety issues that arise from the field safety process, risk management, impending deployment decisions, or any other source. Issues that require the attention of senior management are presented to Waymo's Safety Board.

The Waymo Safety Board brings together executive leaders from our Safety, Engineering and Product teams to resolve these safety issues, approve new safety activities, and ensure our entire safety framework is kept current. This process ensures that the most important safety issues affecting Waymo's AVs are brought to the attention of Waymo's senior management and that those managers are accountable for ensuring that Waymo takes appropriate actions to address those issues.

# Waymo's Readiness Determinations

Waymo uses the safety methodologies described above on a daily basis throughout the life cycle of our AVs. On any given day Waymo may be doing V&V testing on a particular passenger vehicle platform, performing structured tests at a testing facility to confirm certain simulation parameters, conducting public road testing of various vehicle platforms at different levels of hardware and software development, performing hazard analysis on a specific aspect of onboard software, performing detailed statistical analysis to support the assessment of collision risks in a particular ODD, analyzing field reports concerning an operational issue possibly affecting the safety of our AVs or passengers, and studying how to improve one or more of its safety methodologies.

Of course, at certain points in time Waymo needs to make a discrete determination resting on those methodologies with regard to the readiness of a specific configuration of the ADS for a specific deployment. Any decision to use a new version of our ADS, including a software upgrade, a major hardware change, progression to a new operating mode (e.g., a transition from public road testing with trained vehicle operators to driverless deployment), or any significant expansion of our ODD, requires a readiness determination focused on the ODD, the specific use case, and the particular vehicle platform. For example, readiness to operate our AVs in a residential environment with 25 m.p.h. speed limits may require a determination and supporting analysis different from that used in approving operation of the same AV in an environment with a mix of speed limits and traffic conditions.

Each determination to move to the next level in the progression of the ADS begins with application of the methodologies previously described. Determinations to move from public road testing with trained vehicle operators to driverless operations, of course, are conducted at the greatest level of detail. Going completely driverless entails extremely rigorous analysis of expected behaviors and risks within the ODD, including unique risks presented by the absence of a human driver (e.g., responding to system failures through fallback maneuvers that do not rely on human intervention). An important element of the process when moving to driverless operation is purposeful gradualism, so that the scale of the change starts small in terms of the extent of the ODD and volume of operations and gradually ramps up as the ADS proves to be performing as expected.

Accordingly, a determination of an ADS's readiness is not a one-time event.[39] With repeated upgrades of the software, introduction of new vehicle platforms and hardware, and repeated expansion of AV operations into different ODDs, Waymo needs and has an effective decision-making process to support its readiness determinations. In brief, that process entails an ordered examination of the relevant outputs from all of our safety methodologies combined with careful safety and engineering judgment focused on the specific facts relevant for a particular determination.

In preparing to launch a new software version or to expand the ODD, we establish performance criteria for the ADS reflecting operation within a reasonable level of risk in specific areas of performance. Our actual performance criteria in a given ODD are conservative, rate-based, and focus greatest attention on events of higher predicted severity. And, of course, the performance criteria necessarily vary across different ODDs, use cases, and vehicle platforms. Before launching, we evaluate the ADS's likely performance in meeting each of those separate criteria and apply the necessary judgment to reach an aggregate assessment of the ADS's likely safety performance under the intended operating conditions.

The entire process culminates in a determination, based on all of the information derived from our various safety methodologies, as to whether the ADS is in fact ready for testing and/or deployment under the new conditions. One guiding principle is whether use of the new software, expansion of the ODD, removal of the trained vehicle operator or other major change will result in a condition of safety, which in the automotive engineering context is defined as the absence of unreasonable risk.[40] This requires careful analysis of all of the signals provided by our various methodologies rather than a narrow analysis of any single factor as well as sophisticated expert engineering and safety professional judgment to synthesize the information into a well-supported conclusion. If Waymo has not established readiness for the change, that new step is delayed while the critical issues revealed by the readiness analysis are addressed. Waymo will approve when it determines the ADS is ready for the new conditions without creating any unreasonable risks to safety. Constant monitoring and field safety review of a new testing phase or deployment reveals whether the AV's performance is meeting the metrics on which the readiness review and approval were based. Waymo acts quickly to address emergent safety issues that are identified during this monitoring. As the deployment scale increases and the available data grow in volume, performance targets are continuously refined.

---

[39] This is why certain external methodologies that focus on readiness determinations, although worthy of full consideration, are not fully applicable in Waymo's context. Those methodologies may be better suited for AVs designed for sale to the public involving a single platform with a broad ODD. Those AVs will not be owned and controlled by the ADS developer and, other than by software updates, will not be subject to continuous improvement. Waymo's readiness determinations are just as rigorous as those involving products for sale to consumers, but Waymo's purposeful and gradual scaling of any new operation and continued direct control over the AVs makes the assessment of risks, the ability to immediately detect problems, and the opportunity to effect needed changes to ensure rapid remediation of a safety issue quite different from a typical consumer product launch.

[40] As noted in the section above on "Approaches to Assessing Automated Vehicle Safety," the fundamental safety principle of "avoidance of unreasonable risk" is central to many leading methodologies focused on automotive safety, including ISO 26262 and SOTIF.

# Conclusion

Waymo bases its safety activities on this forward-looking philosophy: *Waymo will reduce traffic injuries and fatalities by driving safely and responsibly, and will carefully manage risk as we scale our operations*. Waymo's safety methodologies provide the pillars of an overall safety process that is well designed to bring that philosophy to fruition as Waymo's ADS progresses through development, testing and deployment in a wide range of vehicle platforms, use cases, and ODDs. Waymo will continue to apply and adapt those methodologies, and to learn from the important contributions of others in the AV industry, as we continue to build an ever safer and more able ADS. Waymo will base decisions to deploy our ADS in new environments and under new operating conditions on these strong safety foundations and proceed when these methodologies support a determination of readiness.